\documentclass{article}
\usepackage{spconf,amsmath,graphicx,xspace,xcolor}

\usepackage{hyperref}

\usepackage{xcolor}
\hypersetup{
    colorlinks,
    linkcolor={magenta!80!black},
    citecolor={green!100!},
    urlcolor={magenta!80!black}
}
\usepackage[capitalise]{cleveref}
\usepackage{cleveref}

\newcommand{\myfirstpara}[1]{\noindent \textbf{#1}:}
\newcommand{\mypara}[1]{\vspace{0.5em} \myfirstpara{#1}}

\newcommand{\etal}{\textit{et al.}\xspace}
\newcommand{\Depthformer}{\texttt{Depthformer}\xspace}
\newcommand{\KITTI}{\texttt{KITTI}\xspace}
\newcommand{\NYU}{\texttt{NYUV2}\xspace}
\newcommand{\Transbins}{\texttt{Transbins}\xspace}
\newcommand{\MLP}{\texttt{MLP}\xspace}
\newcommand{\MVT}{\texttt{MVT}\xspace}

\newcommand{\ViT}{\texttt{VIT}\xspace}
\newcommand{\ViTs}{\texttt{VITs}\xspace}

\newcommand{\MVTs}{\texttt{MVT}s\xspace}
\let\OLDthebibliography\thebibliography
\renewcommand\thebibliography[1]{
  \OLDthebibliography{#1}
  \setlength{\parskip}{0pt}
  \setlength{\itemsep}{0pt plus 0.3ex}
}

\usepackage{enumitem}

\title{Depthformer: Multiscale vision transformer for monocular depth estimation with Global Local Information Fusion}
\name{Ashutosh Agarwal \qquad Chetan Arora}
\address{Indian Institute of Technology Delhi}
\begin{document}
%
\maketitle
\begin{abstract}

Attention-based models such as transformers have shown outstanding performance on dense prediction tasks, such as semantic segmentation, owing to their capability of capturing long-range dependency in an image. However, the benefit of transformers for monocular depth prediction has seldom been explored so far. This paper benchmarks various transformer-based models for the depth estimation task on an indoor NYUV2 dataset and an outdoor KITTI dataset. We propose a novel attention-based architecture, \texttt{Depthformer} for monocular depth estimation that uses multi-head self-attention to produce the multiscale feature maps, which are effectively combined by our proposed decoder network. We also propose a \texttt{Transbins} module that divides the depth range into bins whose center value is estimated adaptively per image. The final depth estimated is a linear combination of bin centers for each pixel. Transbins module takes advantage of the global receptive field using the transformer module in the encoding stage. Experimental results on NYUV2 and KITTI depth estimation benchmark demonstrate that our proposed method improves the state-of-the-art by \textbf{3.3\%}, and \textbf{3.3\%} respectively in terms of Root Mean Squared Error (RMSE). Code is available at \href{https://github.com/ashutosh1807/Depthformer.git}{https://github.com/ashutosh1807/Depthformer.git}.
\end{abstract}
\begin{keywords}
depth estimation, transformer, attention, adaptive bins
\end{keywords}

\section{Introduction}
\label{sec:intro}
\begin{figure*}[h]
    \centering
    \includegraphics[scale=0.30]{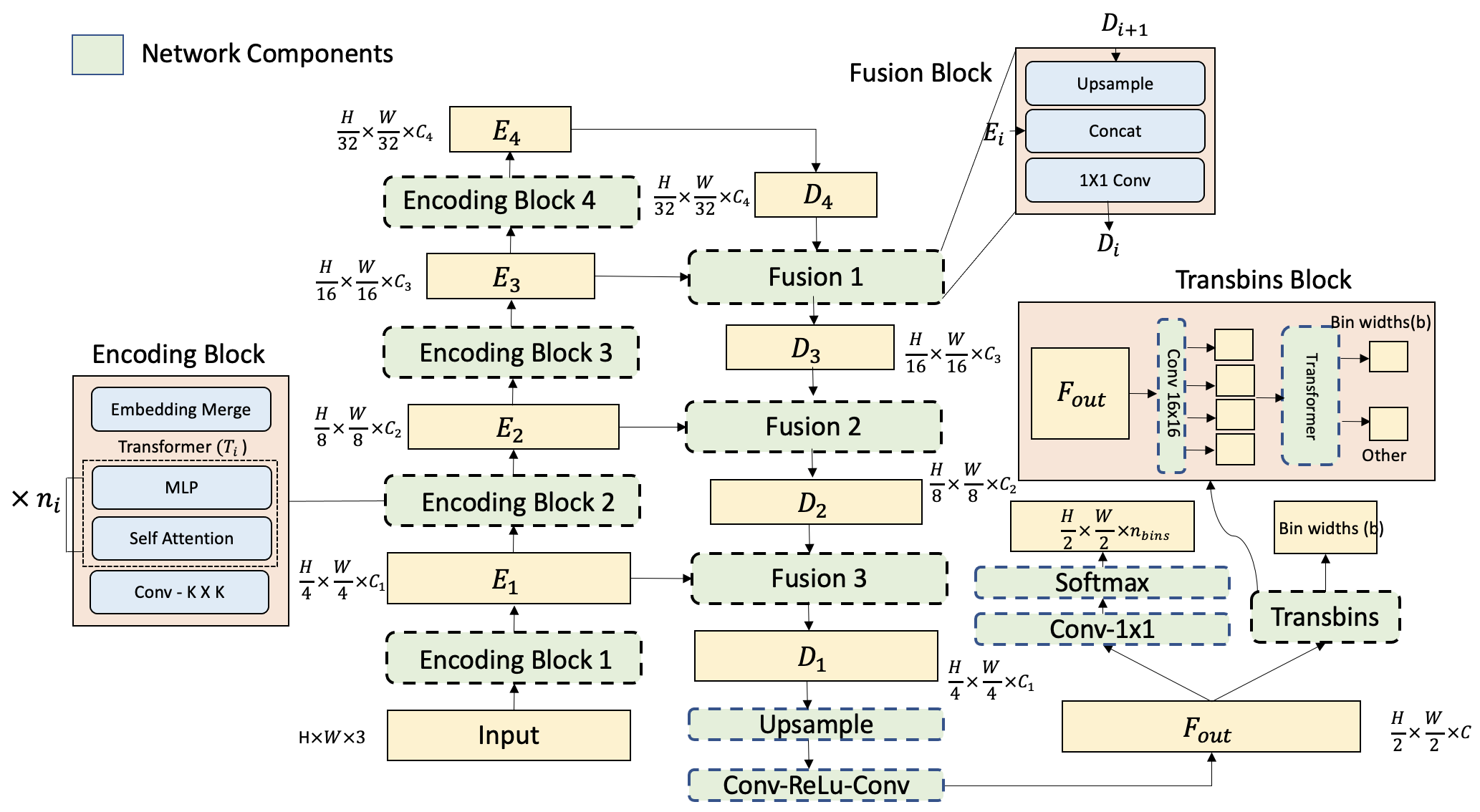}
    \caption{\small \textbf{The architecture of our proposed method, Depthformer} : \MVT produces feature maps of multiple resolutions hierarchically fused by the decoder network to produce output feature map $F_\text{out}$. $F_\text{out}$ is fed to the Transbins module predicts the bin widths. Pixel-wise probability distribution over the bins centers is finally predicted by using a $1 \times 1$ convolution followed by \textit{softmax} activation.  }
    \label{fig:arch}
\end{figure*}

Depth estimation from a single image is fundamental for many applications, from virtual reality to low-cost autonomous driving. Almost all of the current techniques to estimate depth from a single image are based on convolutional neural networks (CNN) with a U-Net-based encoder-decoder architecture~\cite{ eigen, DORN, bts, DAV, chenetal, goddard, Depth2015Liu, Kuznietsov_2017_CVPR, ganetal}. The encoder is typically an image classification network trained on Imagenet~\cite{imagenet}, and the decoder aggregates multiscale features to produce final dense depth. As earlier works have pointed out, the features extracted from a CNN have a local receptive field~\cite{segformer, pvtv1}. For dense prediction tasks such as semantic segmentation and depth estimation, a pixel must have a global receptive field about the scene along with the local information for a more accurate estimation.


The recent success of transformers in Natural Language Processing tasks has created considerable interest among researchers to introduce them in computer vision tasks owing to their capability of capturing long-range dependencies. However, earlier works in Transformer based vision architectures mainly focused on Classification and Object Detection \cite{vit, detr}. With the advent of Multiscale Vision Transformers (\MVTs), researchers have also started using transformer-based architectures as encoders for dense prediction tasks like semantic segmentation~\cite{pvtv1, swin, segformer, twins, mpvit}. Most of the works using \MVTs showcase their dense prediction performance for semantic segmentation. On the contrary, depth estimation is even more difficult for two reasons: (1) It is a continuous prediction task, and (2) It is an ill-posed problem owing to the inherent scale ambiguity. To the best of our knowledge \cite{dense} is the only work that proposes to use vision transformer (\ViT) during the encoding stage. However, it applies \ViT only on feature maps with \textit{1/16th} scale extracted using a CNN architecture. Motivated by these observations, we propose a novel transformer-based encoder for depth estimation that uses multiheaded self-attention to produce hierarchical feature maps. We also benchmark the performance of recently proposed vision transformers (\ViTs) for semantic segmentation on the monocular depth estimation task. 

Using a transformer-based encoder increases the receptive field of the network, but for the dense prediction task, pixels also must understand local information. For example, a pixel must know that it lies on a boundary of an object or that it belongs to a group of pixels on a co-planar surface. Previous works that use \MVTs \cite{segformer} as encoders have a decoder design in which they upscale the encoded features of varying resolutions to a fixed resolution and fuse them using \MLP layers. However, directly upsampling to a higher resolution and fusion results in a loss in local information. Motivated by this, we propose a novel decoder that iteratively upsamples feature maps and fuses them with the encoder features, starting from the lowest resolution and moving towards the high resolution. Iterative upsampling and fusion helps propagate global information to high-resolution local information preserving features. 

In alignment with the state-of-the-art (SoTA) \cite{adabins}, we model the depth estimation task as an ordinal regression problem that discretizes the depth range into a number of intervals or bins adaptively per image. We propose a novel module \Transbins, that takes advantage of fused global information into the feature maps by an \MVT and produces adaptive bins discretized over the depth range.
To demonstrate the efficacy of our proposed network \Depthformer, we conduct extensive experiments on two canonical depth estimation benchmarks, an outdoor driving dataset \KITTI~\cite{kitti} and an indoor dataset \NYU~\cite{nyu}, achieving SoTA on both. 

\mypara{Contributions} 
To summarise, our contributions are - \textbf{(1)} We propose a novel encoder-decoder network that uses self-attention to predict multiscale feature maps that are effectively fused by our decoder. \textbf{(2)} We propose a novel \Transbins module that predicts adaptive bins discretized over the depth range whose centers are used to predict final depth. \textbf{(3)} We benchmark the performance of earlier proposed multiscale vision transformers (\MVTs) on monocular depth estimation. \textbf{(4)} Our network \Depthformer achieves state-of-the-art performance on both outdoor dataset \KITTI and indoor dataset \NYU.

\section{Methodology}
\label{sec:methodology}
\myfirstpara{Problem Definition}
Following \cite{adabins, DORN}, we model depth estimation as an ordinal regression task. Given an input $I$, our network predicts: (1) Bin widths, $b$, that discretize continuous depth range into a number of intervals, $n_\text{bins}$, adaptively per image. (2) Pixel-wise probability distribution over the adaptive bins. The final depth, $d$, at a pixel is the linear combination of the probability scores at the pixel and per image depth-bin-centers.  

\mypara{Architecture}
This section introduces the architecture for the encoder and decoder of our proposed \Depthformer and the training loss that we follow. Our encoder produces the feature maps of varying resolutions using a \MVT given an input image. The decoder then fuses these multiscale features to predict feature map $F_\text{out}$. Next, the proposed \Transbins module uses $F_\text{out}$ as input to predict adaptive bins $b$ discretized over the depth range. The probability distribution over the bins is estimated by applying a convolution operation on $F_\text{out}$ followed by the \textit{softmax} activation. We describe a more detailed architecture below. 

\mypara{Encoder}
\label{sec:enc}
Given an image $I$ of size $H \times W \times 3$, our encoder aims to create feature maps at multiple resolutions using the Transformer framework.
Input $I$ is first convolved with learned $C_1$ kernels of size $K_1$ and a stride of $4$ to produce a feature map of size $\frac{H}{4} \times \frac{W}{4} \times C_1$. This feature map is flattened to produce a sequence of $\frac{HW}{4^2}$ feature vectors, each of dimension $C_1$ that are fed to the transformer block $T_1$. Transformer $T_1$ applies $n_1$ self-attention - \texttt{MLP} layers to produce the output, which is reshaped to a feature map $E_1$ of size $\frac{H}{4} \times \frac{W}{4} \times C_1$. This process is performed repeatedly, with a stride of $2$ for the convolution operation in the following layers. Hence, the encoder produces feature maps ${E_1, E_2, E_3, E_4}$ of resolutions $\{\frac{1}{4}, \frac{1}{8}, \frac{1}{16}, \frac{1}{32}\}$ respectively with channels $[C_1, C_2, C_3, C_4] = \{64, 128, 320, 512\}$ respectively, that are then fed to the decoder. 

Note that, the self attention in the Transformer Block ($T_i$) is optimised by Spatial attention Reduction (\texttt{SRA}) as suggested in \cite{pvtv1}. \texttt{SRA} is implemented via a convolution layer with kernel size $R_i$ and stride $R_i$ to project the key-value pairs, hence resulting in $\frac{n}{R^2}$ compressed key-value pairs where $n$ is the number of input vectors. For the convolution operation with kernel size $K$, stride $S$ and padding $P$, we use $K_1 = 7, S_1 = 4, P_1 = 3$ for the first layer and $K_i = 3, S_i = 2, P_i = 1$ for $i = 2, 3, 4$. A kernel size $K$ greater than the stride $S$, encourages shared information between the adjacent feature vectors to produce a smoother feature maps. The number of self-attention \MLP layers for the four transformer blocks are: $[N_1, N_2, N_3, N_4] = \{3, 8, 27, 3\}$. Spatial reduction ratios for the four transformers are: $[R_1, R_2, R_3, R_4] = \{8, 4, 2, 1\}$. Fig. \ref{fig:arch} gives an overview of our encoder design.

\mypara{Decoder}
Earlier works using \MVT for dense prediction \cite{segformer}  upsample feature maps at varying resolutions to a resolution of $\frac{1}{4}$, and reduce their channel dimensions to $C$ using $1 \times 1$ convolutions. The features are then concatenated and finally fused to predict the output of size $\frac{H}{4} \times \frac{W}{4} \times n_{cls}$ for the segmentation task, where $n_{cls}$ are the number of classes. The feature maps are then upsampled using interpolation to $H \times W \times n_{cls}$ which helps in producing a smoother estimation.
Such a decoder design suffers from the loss of local information due to the smoothing effect of interpolation. Earlier works using CNN have used Feature Pyramid Network (\texttt{FPN}) \cite{fpn} architecture design to preserve the local details. We adopt a similar design and a decoder that iteratively fuses feature maps from the lowest resolution for \MVTs.

Effectively, for encoder feature maps $E_1$, $E_2$, $E_3$, and $E_4$ with resolution $\{\frac{1}{4}, \frac{1}{8}, \frac{1}{16}, \frac{1}{32}\}$, we iteratively perform the following operation
\begin{align}
D_{i} = \texttt{Conv}\{\texttt{Concat}[\texttt{Upsample}(D_{i+1}),E_i]\} \nonumber \\
  i = {1,2,3,4}
\end{align}
This procedure produces a map feature map $F_\text{out}$ of size $\frac{H}{2} \times \frac{W}{2} \times C$ which is fed to the \Transbins module as shown in Fig. \ref{fig:arch}. We select $C = 128$ as in \cite{adabins}. To upsample the feature maps we have used transposed convolution with kernel size $k = 2$ and stride $s = 2$.

\mypara{Transbins}
Adabins \cite{adabins} predicts adaptive bins and attentual maps, and fuse the later with the feature map from decoder $F_\text{out}$. The motivation is to fuse global information in the attenual maps with the decoder features. We take advantage of the encoded global information in $F_\text{out}$ via our encoder to only predict bin widths from the full-scale \ViT. To predict the distribution over the bins, we use $1 \times 1$ convolution over $F_\text{out}$ followed by \textit{softmax} to predict the output of size $\frac{H}{2} \times \frac{W}{2} \times n_{\text{bins}}$.

The final depth is predicted by linear combination over the bin centers as in \cite{adabins}, which is then upsampled using bilinear interpolation to predict the depth at full resolution. The effectiveness of our proposed \Transbins approach against Adabins for bin widths prediction can be seen in Table \ref{tab:dec} which reduces the RMSE error by 14.2 \%. 

\begin{table}
\resizebox{\columnwidth}{!}{%
\begin{tabular}{l|c|c|c|c|c}
\hline
Method          &             $RMSE\downarrow$ & $Rel\downarrow$    & $\delta_1\uparrow$ & $\delta_2\uparrow$ & $\delta_3\uparrow$   \\
\hline
\textit{Eigen \etal \cite{eigen}} & 0.641  & 0.158  & 0.769                      & 0.95                       & 0.988                                                              \\
DORN \cite{DORN}      & 0.509     & 0.115   & 0.828                      & 0.965                      & 0.992                                                                 \\
Chen \etal \cite{chenetal}   &  0.514    & 0.111   & 0.878                      & 0.977                      & 0.994                                                                \\
VNL  \cite{Yin_2019_ICCV}   &   0.416  & 0.108    & 0.875                      & 0.976                      & 0.994                                                               \\
BTS \cite{bts}  & 0.392             & 0.110       & 0.885                      & 0.978                      & 0.994                                                                 \\
DAV      \cite{DAV} & 0.412               & 0.108        & 0.882                      & 0.980                       & 0.996                                              \\
DPT-Hybrid \cite{dense}& \underline{0.357} & 0.110 & 0.904 & \textbf{0.988} & \textbf{0.998}   \\
Adabins \cite{adabins}  &  0.364            & \underline{0.103}              & \underline{0.903}                      & \underline{0.984}                      & \underline{0.997}                                             \\      \hline
\textbf{Depthformer (ours)} & \textbf{0.345} & \textbf{0.100} & \textbf{0.911} & \textbf{0.988}  & \underline{0.997}    \\
\hline
\end{tabular}
}
\caption{Results on NYUV2 Dataset. The best results are in \textbf{bold} and second best are \underline{underlined}. Our method outperforms the previous SoTA methodsin most of the metrics. }
\label{tab:nyu}

\end{table}





\mypara{Training loss}
We train our network on a sum of scaled version of the Scale-Invariant (SI)  loss introduced by Eigen \etal \cite{eigen} $L_{\text{SILog}}$ and Chamfer loss $L_{\text{Chamfer}} $ \cite{adabins}. $L_{\text{SILog}}$ reduces the difference between the predicted depth map and the ground truth depth map. $L_{\text{Chamfer}} $ encourages  the bin centers to be close to the actual ground truth depth values and vice versa.
%
%
%
%
\begin{equation}
L_{\text{total}} = L_{\text{SILog}} + \gamma L_{\text{Chamfer}}.
\end{equation}
We have used $\gamma =  0.1$ as in \cite{adabins}.

\section{Experiments}
\label{sec:experiments}

\begin{figure*}[h]
    \centering
    \includegraphics[scale=0.25]{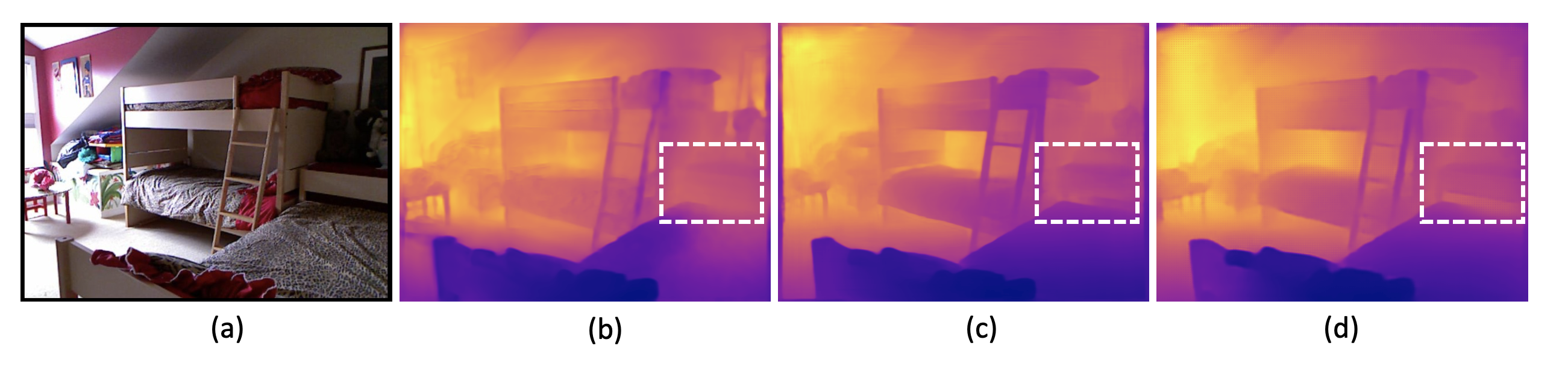}
    \caption{\small Visualisation of predicted Depth map for input image (a) for (b) Adabins \cite{adabins} (c) DPT-Hybrid \cite{dense} (d) \textbf{Depthformer (Ours)}. Our method is able to predict a more accurate depth estimation owing to it's capabilities of capturing long range information.}
    \label{fig:vis}
\end{figure*}

\mypara{Datasets} 
\textbf{NYU Depth v2} \cite{nyu} contains indoor scenes dataset with 640 × 480 resolution images and depth maps with the upper bound of 10 m. We have trained our network on a 24K subset of dataset with a random crop of 576 × 448. We evaluate on the pre-defined center cropping by Eigen \etal \cite{eigen} on test set of 654 images. 

\textbf{KITTI} \cite{kitti} is an outdoor scenes dataset containing stereo images with 1241 × 376 resolution and corresponding 3D laser scans of outdoor scenes. We train our network on a subset of around 26K images from the left view with a random crop of size 704 × 352 and test on 697 images. To compare our performance, we evaluate on a predefined crop by Garg \etal \cite{garg} on 697 test images with a maximum value of 80 m.

\mypara{Evaluation Metrics} 
We use the standard five metrics used in earlier works~\cite{eigen} to compare our method against SoTA. Given the predicted depth $d_p$, and the ground truth depth $d^*_p$ at a pixel $p$, and $n$ denoting the total number of pixels in an image, the error metrics are defined as: \\ \textbf{Root mean squared error (RMSE):} \begin{small} 
$\sqrt{\left.\frac{1}{n} \sum_{p = 1}^{n}\left(d_{p}-d^{*}_{p}\right)^{2}\right)}$
\end{small}  
\textbf{Average relative error (REL):} $\frac{1}{n} \sum_{p = 1}^{n} \frac{\left|d_{p}-d^{*}_{p}\right|}{d^{*}_{p}}$ \\ 
\textbf{Threshold accuracy ($\delta_{i}$) :} 
$\%$ of $d_{p}$ such that $\max \left(\frac{d_{p}}{d^{*}_{p}}, \frac{d^{*}_{p}}{d_{p}}\right)=\delta_i < \text{Thr}$, where $\text{Thr} = 1.25,1.25^2,1.25^3$.  

\mypara{Implementation Details} 
To train our network, we use the AdamW optimizer \cite{adamw} with weight-decay $0.1$. We follow the training methodology prescribed in \cite{onecycle}, and use 1-cycle policy for the learning rate with \textit{max\_lr} = $10^{- 4}$, linear warm-up from \textit{(3/10)max\_lr} to \textit{max\_lr} for the first 50\% of iterations, followed by cosine annealing to \textit{(3/10)max\_lr}. The network has been trained on 4 V100 GPUs with a 32GB memory, with a batch size of 16 for both \NYU and \KITTI. 

\myfirstpara{Comparison with State of the Art}
Table \ref{tab:nyu} and \ref{tab:kitti} demonstrate the performance of our proposed method, \Depthformer, with the previous SoTA methods. We consistently outperform across all the metrics for both outdoor scenes dataset \KITTI and indoor dataset \NYU. Our method reduces the $RMSE$ by 3.5\% for \NYU against SoTA DPT-Hybrid \cite{dense} and by 3.2\% against Adabins \cite{adabins}. As shown in Fig. \ref{fig:vis}, our method Depthformer predicts a more accurate depth image in comparision to SoTAs \cite{dense} and \cite{adabins}.

\begin{table}
\resizebox{\columnwidth}{!}{%
\begin{tabular}{l|c|c|c|c|c}
\hline
Method             & $RMSE\downarrow$   & $Rel\downarrow$              & $\delta_1\uparrow$ & $\delta_2\uparrow$ & $\delta_3\uparrow$   \\
\hline
Eigen \etal \cite{eigen}  & 6.307   & 0.203   & 0.702                      & 0.898                      & 0.967                                                              \\
Goddard \etal \cite{goddard}  & 4.935   & 0.114     & 0.861                      & 0.949                      & 0.976                                                           \\
Gan \etal  \cite{ganetal}  & 3.933     & 0.098      & 0.890                       & 0.964                      & 0.985                                                          \\
DORN \cite{DORN}   & 2.727      & 0.072      & 0.932                      & 0.984                      & 0.994                                                          \\
Yin \etal  \cite{Yin_2019_ICCV}  & 3.258    & 0.072      & 0.938                      & 0.990                      & \underline{0.998}                                                           \\
BTS      \cite{bts}   & 2.756           & \underline{0.059}                & 0.956                      & 0.993                      & \underline{0.998}                                                          \\

DPT-Hybrid \cite{dense}& 2.573 & 0.062  & 0.959 & 0.995 & \textbf{0.999}   \\
Adabins   \cite{adabins}       & \underline{2.360}       & \textbf{0.058}           & \underline{0.964}                      & \underline{0.995}                      & \textbf{0.999}                                               \\
\hline
\textbf{Depthformer (ours)} & \textbf{2.285}  & \textbf{0.058}  & \textbf{0.967} & \textbf{0.996} & \textbf{0.999}  \\
\hline
\end{tabular}}
\caption{Results on \KITTI Dataset. The best results are in \textbf{bold} and second best are \underline{underlined}.  }
\label{tab:kitti}

\end{table}

\mypara{Benchmarking of \ViTs on Monocular depth estimation} 
Table \ref{tab:benchmark} shows the performance of various SoTA \ViTs for monocular depth estimation on the \NYU dataset. For a fair comparison, we have taken model variants with a similar number of trainable parameters. For this experiment, we have taken the decoder as in \cite{segformer}. All the \ViTs outperform the CNN-based Resnet-50~\cite{resnet}. MiT-B2~\cite{segformer} shows the best performance 
in comparison to other \MVTs. This can be attributed to shared information across feature vectors as mentioned in Section \ref{sec:enc}. In this work, we initialise the encoder via pretrained weights from MiT-B4~\cite{segformer}. We hope this benchmark can be used as a baseline for future research in depth estimation.

\begin{table}
\centering
\resizebox{\columnwidth}{!}{%
\begin{tabular}{l|c|c|c|c}
\hline
Method    & Reference & Param. (M)& $RMSE\downarrow$ & $REL\downarrow$ \\
\hline
ResNet-50 \cite{resnet}   & \textit{CVPR'16} & 23.5 & 0.510    &    0.152   \\ 
PVTv1 \cite{pvtv1}    &  \textit{ICCV'21}  & 23.9 & 0.508  &    0.166    \\

Swin-T \cite{swin}     &  \textit{ICCV'21} & 27.5 & 0.456    &   0.142     \\
Twins-SVT-S \cite{twins}     & \textit{NeurIPS'21} & 23.5   & 0.443  &     0.141   \\
MiT-B2 \cite{segformer} & \textit{NeurIPS'21}  & 24.2 & \textbf{0.394} &     \textbf{0.118}   \\

MPVIT \cite{mpvit}     &  \textit{CVPR'22} & 22.6 & 0.403 &  0.120     \\
\hline
\end{tabular}}
\caption{Performance of different state of the art multiscale-vision transformers for monocular depth estimation on the benchmark \NYU dataset.}
\label{tab:benchmark}
\end{table}

\mypara{Ablation for Decoder architecture} 
Table \ref{tab:dec} showcases the efficacy of our proposed decoder network. For all these experiments, we have taken the proposed encoding backbone. The baseline uses a decoder network as proposed in \cite{segformer}, which predicts the depth at a resolution of $1/4th$ scale and upsamples it using bilinear interpolation. We first deploy Global Average Pooling (GAP) on top of our decoder to predict adaptive bins. As shown in the table, adding GAP gives a significant improvement over the baseline.
Next, we ablate on our decoder design using the adaptive binning strategy proposed by Adabins \cite{adabins} and our proposed \Transbins. As shown, \Transbins outperforms Adabins by a good margin which validates our claim that encoding global information during the initial layers is a better strategy in comparison to encoding the information at the final layers.

\begin{table}[]
\centering
\begin{tabular}{l|c|c}
\hline
Method                    & $RMSE\downarrow$ & $REL\downarrow$ \\
\hline
Decoder Xie \etal (\cite{segformer})                &   0.375   &     0.114    \\
Decoder(ours) + GAP               &    0.350  &         0.105 \\

Decoder(ours) + Adabins \cite{adabins}               &    0.394  &         0.115 \\

Decoder(ours) + Transbins &   \textbf{0.345}   &   \textbf{0.100}     \\
\hline
\end{tabular}
\caption{Ablation study on \NYU dataset for different decoder designs in the proposed \Depthformer model.}
\label{tab:dec}
\end{table}


\section{Conclusion}
\label{sec:conclusion}

This paper introduces a multiscale vision transformer-based monocular depth estimation technique that achieves state-of-the-art results on \KITTI and \NYU datasets. We also benchmark SoTA transformer architectures for monocular depth estimation task. We hope it motivates researchers to design new architectures to address task specific intricacies.

\mypara{Acknowledgement} We acknowledge National Supercomputing Mission (NSM) for providing computing resources of 'PARAM Siddhi-AI', under National PARAM Supercomputing Facility, C-DAC Pune, and supported by the Ministry of Electronics and Information Technology and Department of Science and Technology, Government of India. This work has also been partly supported by the funding received from DST through the IMPRINT program (IMP/2019/000250).


\bibliographystyle{IEEEbib}
\bibliography{refs}

\end{document}